# Out-of-Sample Extension for Dimensionality Reduction of Noisy Time Series

Hamid Dadkhahi, Marco F. Duarte, *Senior Member, IEEE*, and Benjamin M. Marlin

*Abstract*—This paper proposes an out-of-sample extension framework for a global manifold learning algorithm (Isomap) that uses temporal information in out-of-sample points in order to make the embedding more robust to noise and artifacts. Given a set of noise-free training data and its embedding, the proposed framework extends the embedding for a noisy time series. This is achieved by adding a spatio-temporal compactness term to the optimization objective of the embedding. To the best of our knowledge, this is the first method for out-of-sample extension of manifold embeddings that leverages timing information available for the extension set. Experimental results demonstrate that our out-of-sample extension algorithm renders a more robust and accurate embedding of sequentially ordered image data in the presence of various noise and artifacts when compared to other timing-aware embeddings. Additionally, we show that an out-of-sample extension framework based on the proposed algorithm outperforms the state of the art in eye-gaze estimation.

## I. INTRODUCTION

Recent advances in sensing technology have enabled a massive increase in the dimensionality of data captured from digital sensing systems. Naturally, the high dimensionality of the data affects various stages of the system, from acquisition to processing and analysis of the data. To meet communication, computation, and storage constraints, in many applications one seeks a low-dimensional embedding of the high-dimensional data that shrinks the size of the data representation while retaining the information we are interested in capturing. This problem of *dimensionality reduction* has attracted significant attention in the signal processing and machine learning communities.

The traditional method for dimensionality reduction is *principal component analysis* (PCA) [2], which successfully captures the structure of datasets that are well approximated by a linear subspace. However, in many applications, the data can be best modeled by a *nonlinear manifold* whose geometry cannot be captured by PCA. Manifolds

This work was partially supported by the National Science Foundation under grants IIS-1239341, IIS-1319585, and IIS-1350522. A portion of this work appeared at the IEEE International Workshop on Machine Learning for Signal Processing, 2015 [1].

H. Dadkhahi is with the Electrical and Computer Engineering (ECE) Department and the College of Information and Computer Sciences (CICS), University of Massachusetts, Amherst. E-mail: hdadkhahi@cs.umass.edu. M. F. Duarte is with the ECE Department, University of Massachusetts, Amherst, Email: mduarte@ecs.umass.edu. B. Marlin is with CICS, University of Massachusetts, Amherst, E-mail: marlin@cs.umass.edu.

are low-dimensional geometric structures that reside in a high-dimensional ambient space despite possessing merely a few degrees of freedom. Manifold models are a good match for datasets associated with a physical system or event governed by a few continuous-valued parameters. Once the manifold model is formulated, any point on the manifold can be essentially represented by a low-dimensional vector. *Manifold learning methods* aim to obtain a suitable nonlinear embedding into a low-dimensional space that preserves the local structure present in the higher-dimensional data in order to simplify data visualization and exploratory data analysis. Examples of manifold learning methods include Isomap [3], locally linear embedding [4], Laplacian eigenmaps [5], and Hessian eigenmaps [6].

All the aforementioned dimensionality reduction methods assume that the data points are stationary and independent. However, in many real-world applications in vision and imaging we encounter time series data. In recent years, several attempts have been made to take advantage of temporal information of the data in order to discover the dynamics underlying the manifold. Spatio-temporal Isomap (ST-Isomap) [7] empirically alters the original weights in the graph of local neighbors to emphasize similarities among temporally related points. Similarly, the *temporal Laplacian eigenmap* [8] incorporates temporal information into the Laplacian eigenmap (LE) framework. Furthermore, [9] extends the global coordination model of [10] to a dynamic Bayesian network (DBN) by adding links among the intrinsic coordinates to account for temporal dependency. In addition, [11] applies a semi-supervised regression model to learn a nonlinear mapping from temporal data. Finally, temporal information has been similarly used in other image processing tasks to improve learning performance, e.g., [12], and its use is natural when processing video sequences.

When the data to be embedded corresponds to a time series, all these temporal frameworks of dimensionality reduction generate better quality embedded spaces than the initial approaches in the sense that the embedding indicates the dynamics of the data. However, all of the mentioned methods are designed only for a given set of training points, with no straightforward extension of these timing-aware embeddings for out-of-sample points.

On the other hand, many of these dimensionality reduction algorithms (and Isomap in particular) are sensitive to noise and artifacts [13]. This is not desirable for real-world applications since the data are usually contaminated with noise and various artifacts due to imperfect sensors or human mistakes. This sensitivity is particularly relevant for emerging architectures for very low power sensing, which are subject to increased presence of noise and artifacts [14].

In this paper, we address the problem of out-of-sample extension of nonlinear manifold embeddings for the specific case where the points to be extended correspond to a time series. The aim is to use the sequential ordering of the data points to improve the resilience of the embedding of out-of-sample points to various noise and artifact models. This improvement is achieved by enforcing a compactness for pairs of points within a temporal neighborhood window. Our focus in this paper is on Isomap for the sake of concreteness, but our formulation is generic enough that it can also be applied to additional nonlinear manifold learning algorithms that are formulated as optimization problems.

The contributions of this paper can be summarized as follows. We propose an optimization problem for out-of-sample manifold extension with a cost function that incorporates two distinct terms. The role of the first term is to preserve the manifold structure using only the spatial information from all data points (original and out-of-sample). The second term, referred to as spatio-temporal compactness, aims to leverage temporal information among out-of-sample points. The compactness term is defined based on a weighted Laplacian for the temporal neighborhood graph of the out-of-sample points. The use of temporal information is suitable when processing sequences of data points that correspond to the traversal of continuous paths in the manifold, which is commonly observed in video sequences whose frames traverse paths in image manifolds. Similarly, our use of temporal information in out-of-sample extension is motivated by the application of manifold modeling of video sequence frames, where a reference manifold is learned from a clean data set. Existing methods that aim to leverage spatio-temporal information in manifold learning are not compatible with this application due to the lack of temporal information on the reference/clean data set. We showcase the benefit of leveraging spatio-temporal information for out-of-sample extension by demonstrating higher resilience to the presence of noise and distortion in the points involved in the extension. To the best of our knowledge, the idea of introducing the spatio-temporal compactness for manifold models, its particular setup for the out-of-sample extension problem, and the proposed optimization formulation to solve the latter problem, are novel to our work.

The application scenario of our proposed scheme is as follows. In a training stage for nonlinear manifold learning, we have full control of the environment, which enables us to capture noise-free data. We then use the captured training data to learn the underlying manifold. In the testing stage (which corresponds to the normal operation of the sensors), we capture data points of lower quality, possibly contaminated with different artifacts/noise patterns, in a sequential order (i.e., as a time series). We then extend the original nonlinear manifold embedding to the newly acquired (noisy) samples using our proposed algorithm while leveraging their timing information.

The rest of this paper is organized as follows. We first review the literature on manifold learning, in particular Isomap, and its spatio-temporal, denoising, and out-of-sample extensions in Section II. In Section III, we derive a framework for out-of-sample extension of Isomap in which temporal information of the data points is taken into account. Experimental results on a real-world dataset is presented in Section IV. We conclude this paper with discussions on the proposed algorithm and future work in Section VI. We also summarize key notation used through the paper in Table I.

## II. BACKGROUND

### A. Manifold Models and Manifold Learning

A set of data points $\mathcal{U} = \{u_1, u_2, \ldots, u_n\}$ in a high-dimensional ambient space $\mathbb{R}^d$ that have been generated by an event featuring $m$ degrees of freedom correspond to a sampling of an $m$-dimensional *manifold* $\mathcal{M} \subset \mathbb{R}^d$. Given the high-dimensional dataset $\mathcal{U}$, we would like to find the parameterization that has generated the manifold. One way to discover this parametrization is to embed the high-dimensional data on the manifold to a low-dimensional space $\mathbb{R}^m$ so that the local geometry of the manifold is preserved. *Dimensionality reduction methods* are devised so as to preserve such geometry, which is measured by a neighborhood-preserving criteria that varies depending on the specific algorithm.

*Principal component analysis* (PCA) is perhaps the most popular scheme for linear dimensionality reduction of high-dimensional data [2]. PCA is defined as the orthogonal projection of the data onto a linear subspace of lower dimension $m$ such that the variance of the projected data is maximized. Unfortunately, PCA fails to preserve the geometric structure of a *nonlinear manifold*, i.e., a manifold where the mapping from parameters to data is nonlinear. Several *nonlinear manifold learning methods* can successfully embed the data into a low-dimensional model while preserving such local geometry in order to simplify the parameter estimation process.

### B. Isomap

For nonlinear manifold learning, this paper focuses on the *Isomap* algorithm, which aims to preserve the pairwise *geodesic distances* between data points [3]. The geodesic distance is defined as the length of the shortest path between two data points $u_i$ and $u_j$ ($u_i, u_j \in \mathcal{M}$) along the surface of the manifold $\mathcal{M}$ and is denoted by $d_G(u_i, u_j)$. Isomap first finds an approximation to the geodesic distances between each pair of data points in a sampling of the manifold $\mathcal{U}$ by constructing a neighborhood graph in which each point is connected only to its $k$ nearest neighbors; the edge weights are equal to the corresponding pairwise distances. For neighboring pairs of data points, the Euclidean distance provides a good approximation for the geodesic distance, i.e.,

$$d_G(u_i, u_j) \approx \|u_i - u_j\|_2 \quad \text{for} \quad u_j \in \mathcal{N}_k(u_i), \quad (1)$$

TABLE I: List of important notations.

| Notation | Description | Notation | Description |
|---|---|---|---|
| $\mathcal{U}$ | set of training points | $n$ | cardinality of $\mathcal{U}$ |
| $\mathcal{L}$ | embedding set for training points $\mathcal{U}$ | $N$ | cardinality of $\mathcal{Y}$ |
| $\mathcal{Y}$ | set of out-of-sample points | $m$ | dimensionality of $\mathcal{L}$ (and $\mathcal{X}$) |
| $\mathcal{X}$ | embedding set for extension points $\mathcal{Y}$ | $K$ | number of temporal neighbors in $\mathcal{T}$ |
| $\mathcal{T}_K$ | temporal graph for $\mathcal{Y}$ | $\alpha$ | decay parameter for weighting function |
| $\Delta_n$ | squared distances between training points $\mathcal{U}$ | $\lambda$ | regularization parameter |
| $\Delta_X$ | squared distances between training points $\mathcal{U}$ and extension points $\mathcal{Y}$ | $\omega$ | temporal weighting function |

where $\mathcal{N}_k(u_i)$ designates the set of $k$ nearest neighbors in $\mathcal{U}$ to the point $u_i \in \mathcal{U}$. For non-neighboring points, the geodesic distance is estimated as the length of the shortest path along the neighborhood graph, which can be found via Dijkstra's algorithm. Then, multidimensional scaling (MDS) [15] is applied to the resulting geodesic distance matrix to find a set of low-dimensional points that best match such distances.

### C. Out-of-Sample Extension for Isomap

Out-of-sample extension (OoSE) generalizes the result of the nonlinear manifold embedding for new data points. Suppose we have $n$ training data points $\mathcal{U} = \{u_1, u_2, \ldots, u_n\}$. The embedding $\mathcal{L} = \{\ell_1, \ell_2, \ldots, \ell_n\}$ of the training points is obtained by applying MDS to the geodesic distance matrix of the $n$ training points. More precisely, the centralized squared geodesic distances can be obtained as $\Delta_c = -\frac{1}{2}H_n\Delta_nH_n$, where $\Delta_n$ is the matrix of squared geodesic distances (i.e., $\Delta_n[i,j] = d_G^2(u_i, u_j)$) and $H_n$ is the centering matrix defined by the formula $H_n[i,j] = \delta_{ij} - \frac{1}{n}$. Then, the $m$-dimensional embedding $\mathcal{L}$ for the $n$ training points $\mathcal{U} = \{u_1, u_2, \ldots, u_n\}$ is given as the columns of the matrix

$$L = \begin{bmatrix} \sqrt{\lambda_1} \cdot v_1^T \\ \sqrt{\lambda_2} \cdot v_2^T \\ \vdots \\ \sqrt{\lambda_m} \cdot v_m^T \end{bmatrix},$$

where $\lambda_i$ and $v_i$ are the eigenvalues and eigenvectors of the centralized squared geodesic distance matrix $\Delta_c$, respectively.

To obtain the extension of the embedding $\mathcal{L}$ to the out-of-sample (testing) set $\mathcal{Y} = \{y_1, y_2, \ldots, y_N\}$, let $\Delta_X$ denote the $n \times N$ matrix of squared geodesic distances between the $N$ out-of-sample (testing) points $\mathcal{Y}$ and the $n$ training points $\mathcal{U}$; these geodesic distances are estimated as before with an augmented graph including links to the $k$ nearest neighbors of each out of sample point. The out-of-sample extension $\mathcal{X} = \{x_1, x_2, \ldots, x_N\}$ of the embedding $\mathcal{L}$ to the points $\mathcal{Y}$ is given by the columns of the matrix [16], [17]

$$X = \frac{1}{2} L^{\#} (\bar{\Delta}_n \mathbf{1}_N^T - \Delta_X), \qquad (2)$$

where $\mathbf{1}_N$ denotes an all-ones vector of length $N$, $\bar{\Delta}_n$ is the column mean of $\Delta_n$, i.e., $\bar{\Delta}_n = \frac{1}{n}\Delta_n\mathbf{1}_n$, and $L^{\#}$ is the pseudo-inverse transpose of $L$, given by

$$L^{\#} = (L^T)^{\dagger} = \begin{bmatrix} v_1^T / \sqrt{\lambda_1} \\ v_2^T / \sqrt{\lambda_2} \\ \vdots \\ v_m^T / \sqrt{\lambda_m} \end{bmatrix}.$$

Note that, to the best of our knowledge, the literature on out-of-sample extension does not exploit the sequential ordering of data to mitigate the possibility of embedding errors due to the presence of noisy and contaminated data.

### D. Spatio-Temporal Isomap

In Isomap, the neighborhood graph is formed by linking each point in $\mathcal{U}$ to its $k$-nearest neighbors in the same set. *Spatio-Temporal Isomap* (ST-Isomap) leverages sequence/timing information $\{(u_i, t_i)\}$ present for each of the training points $u_i \in \mathcal{U}$. ST-Isomap appends edges between pairs of *adjacent temporal neighbors* (ATN), i.e., pairs of immediate temporal neighbors where $t_j = t_i \pm 1$, to the neighborhood graph. The weights (i.e., distances) between ATNs are scaled down by a factor given by the constant parameter $c_{ATN}$, which is used to emphasize the correlation between a point and its adjacent temporal neighbors.

ST-Isomap also modifies the graph weights of a subset of the $k$-nearest neighbors of each point that satisfy certain spatio-temporal conditions. First, the set of points in a temporal window of size $\epsilon$ around each point $u_i$ is considered as its *trivial matches*, denoted by $\mathcal{T}_\epsilon(u_i)$. Suppose that the point $u_j \in \mathcal{T}_\epsilon(u_i)$ is the closest trivial match to point $u_i$, i.e.,

$$d_G(u_i, u_j) = \min_{u_k \in \mathcal{T}_\epsilon(u_i)} d_G(u_i, u_k).$$

Now, the subset of $k$-nearest neighbors with distances less than or equal to $d_G(u_i, u_j)$ from $u_i$ are considered as *non-trivial matches* and the resulting set is referred to as *common temporal neighbors* (CTN):

$$\text{CTN}(u_i) = \{u_j \in \mathcal{U} : u_j \in \mathcal{N}_k(u_i),$$
$$d_G(u_i, u_j) \le \min_{u_k \in \mathcal{T}_\epsilon(u_i)} d_G(u_i, u_k)\}.$$

In other words, CTN is a subset of the (spatial) nearest neighbor graph, consisting of points which are spatially closer to a given point than its closest temporal neighbor (i.e. the closest trivial match). In a sense, CTN are used to identify data points in the local spatial neighborhood of each point $u_i$ that are more likely to be analogous to $u_i$. Finally, the constant parameter $c_{CTN}$ is used to emphasize the similarity between each point and its CTN set via reducing the corresponding weights by a scaling factor of $c_{CTN}$.

Note that ST-Isomap is devised to better uncover the spatio-temporal structure underlying the manifold data, rather than to faithfully recover the embedding of data contaminated by noise or artifact models. In addition, ST-Isomap does not address the out-of-sample extension problem. Nonetheless, we formulate a straightforward adaptation of out-of-sample extension from Isomap to ST-Isomap in Section IV.

### E. Manifold Denoising

The majority of the literature on manifold models for noisy data focuses on a given set of noisy training data, aiming at reducing the error of the recovered data points in the ambient space rather than the embedded trajectory. In [18] a preprocessing procedure is proposed that enables Isomap and LLE algorithms to achieve more robust reconstruction of nonlinear manifolds in the presence of Gaussian noise. This is achieved by combining locally smoothed values of data points, which are in turn computed via the so-called linear error-in-variables model of local structure. [19] tackles the same problem in Laplacian eigenmaps by reversing the diffusion process employed therein. This is done by approximating the generator of the diffusion process by the graph Laplacian of a random neighborhood graph. [20] proposes a manifold denoising algorithm based on Gaussian process latent variable models (GPLVM) to handle data contaminated by Gaussian and salt-and-pepper noise. GPLVM uses the *maximum a posteriori* estimate of the transformation matrix in the latent variable model to map the latent variables back to the data space. In [21], the authors introduce Locally Linear Denoising (LLD), an algorithm that approximates manifolds with locally linear patches by constructing nearest neighbor graphs. Each point is then locally denoised within its neighborhood. The latter algorithm has been used in denoising of images in the ambient space and improves the performance of classification in the embedded space in the presence of salt-and-pepper, Gaussian, motion blur, and occlusion artifacts.

In [22], a preprocessing procedure, called Manifold Blurring Mean Shift (MBMS), is proposed for denoising manifold data based on blurring mean-shift updates. Each iteration of the MBMS algorithm has two steps. The first step is a blurring mean-shift update that moves data points to the kernel average of their neighbors with a Gaussian kernel of width $\sigma$. Then, a projective step is computed using the local PCA of dimensionality $L$ on the $k$ nearest neighbors of each point.

Note that none of the mentioned algorithms take advantage of temporal information among the images. Moreover, none of these algorithms address the out-of-sample extension problem in the presence of noise and artifacts.

## III. MANIFOLD EXTENSION FOR TIME SERIES

In this section, we describe the formulation of a new algorithm for out-of-sample extension of Isomap nonlinear manifold embeddings that also aims to leverage temporal information in order to improve the quality of the embedding of noisy and corrupted data time series.

Recall that the embedding of out-of-sample points is given by the matrix equation (2). Alternatively, the embedding $\mathcal{X} = \{x_1, x_2, \ldots, x_N\}$ of out-of-sample points $\mathcal{Y} = \{y_1, y_2, \ldots, y_N\}$ can be expressed as the solution to the following optimization problem:

$$X = \arg\min_X \; \left\| \frac{1}{2}\left(\bar{\Delta}_n \mathbf{1}_N^T - \Delta_X\right) - L^T X \right\|_F^2, \quad (3)$$

where $\|.\|_F$ denotes the Frobenius matrix norm, and the columns of the matrix $X$ correspond with the embeddings in $\mathcal{X}$. In our problem setup, we assume that the out-of-sample points $\{y_1, y_2, \ldots, y_N\}$ have been sampled at time instances $\{t_1, t_2, \ldots, t_N\}$, respectively.

To characterize the spatio-temporal similarity, we first define the *spatio-temporal compactness function* $C_\omega(\mathcal{Y}, \tau)$ of geodesic distances and temporal information for the time series:

$$C_\omega(\mathcal{Y}, \tau) := \sum_{i=1}^N \sum_{y_j \in \mathcal{T}_K(y_i)} d_G^2(y_i, y_j) \cdot \omega(\tau_{ij}), \quad (4)$$

where $\omega$ is a temporal weighting function, $\tau_{i,j} = |t_i - t_j|$, and $\mathcal{T}_K(y_i)$ is the set of $K$ nearest temporal neighbors to point $y_i$ in $\mathcal{Y}$: $\mathcal{T}_K(y_i) = \{y_j : j \in \mathbb{N}, \max\{0, i - \lfloor K/2 \rfloor\} \leq j \leq \min\{i + \lfloor K/2 \rfloor, N\}\}$. Note that $K \neq k$ from (1) in general, and there could be many choices for the temporal weighting function. One such choice is the exponential decay function $\omega(\tau_{ij}) = \exp(-\alpha \cdot \tau_{ij})$ with a decay parameter of $\alpha$. This choice of the weighting function is natural since the distance of each point to its immediate temporal neighbors is more likely to be small and this likelihood decreases gradually as we consider subsequent temporal neighbors.

We assume that the geodesic distances in the ambient space of each temporal neighborhood of the out-of-sample data points should be small so that the compactness term in (4) is also small. We aim to incorporate this compactness in the embedding framework. This can be achieved by leveraging the fact that the Euclidean distances in the embedding space are matched to their geodesic counterparts in the ambient space. Therefore, we modify the expression for the compactness by leveraging the expected relationship between the original geodesic distances for $\mathcal{Y}$ and the Euclidean distances of the embedded data points $X$:

$$d_G^2(y_i, y_j) \approx \|x_i - x_j\|_2^2$$
$$= \text{Tr}(B_{ij}X^TX),$$

where $B_{ij}$ is an all zeros matrix of size $N \times N$ except for four elements, $B_{ij}[i, i] = B_{ij}[j, j] = 1$ and $B_{ij}[i, j] = B_{ij}[j, i] = -1$. The compactness term in (4) can then be reformulated as

$$C_\omega(X, \tau) = \sum_{i=1}^N \sum_{y_j \in \mathcal{T}_K(y_i)} \left(\text{Tr}(B_{ij}X^TX) \cdot \omega_{ij}\right),$$

where $\omega_{ij} = \omega(\tau_{ij})$.

In order to incorporate the temporal information among the sequence of out-of-sample points, we introduce the spatio-temporal compactness constraint to the optimization as follows:

$$X = \underset{X}{\arg\min} \ \left\| \frac{1}{2}\left(\bar{\Delta}_n \mathbf{1}_N^T - \Delta_X\right) - L^T X \right\|_F^2 \quad (5)$$
$$\text{subject to:} \quad \mathbf{C}_\omega(X, \tau) < \mu,$$

where $\mathbf{C}_\omega(X, \tau)$ is the *spatio-temporal compactness* among out-of-sample points and $\mu$ is a constant. In words, we would like the spatio-temporal compactness among successive points to be less than a constant $\mu$ in order to leverage the temporal information of data points in deriving the embedding.

Note that we can rewrite the compactness as follows:

$$\mathbf{C}_\omega(X, \tau) = \text{Tr}\left(\sum_{i=1}^{N}\sum_{y_j \in \mathcal{T}_K(y_i)} \omega_{ij} B_{ij} X^T X\right)$$
$$= \text{Tr}\left(A X^T X\right), \quad (6)$$

where the matrix $A$ is given by

$$A = \sum_{i=1}^{N}\sum_{y_j \in \mathcal{T}_K(y_i)} \omega_{ij} B_{ij}.$$

Note that we can alternatively view $A$ as the Laplacian matrix for a weighted temporal neighborhood graph.

Letting $Q = \frac{1}{2}(\bar{\Delta}_n \mathbf{1}_N^T - \Delta_X)$, the optimization in (5) becomes

$$X = \underset{X}{\arg\min} \|Q - L^T X\|_F^2 \quad (7)$$
$$\text{subject to:} \quad \mathbf{C}_\omega(X, \tau) < \mu.$$

We can rewrite the optimization in (7) using a Lagrangian relaxation $(\lambda > 0)$ in the following way:

$$X = \arg\min_X \|Q - L^T X\|_F^2 + \lambda \cdot \mathbf{C}_\omega(X, \tau). \quad (8)$$

Note that $\|A\|_F^2 = \text{Tr}(A^T A)$, where $\text{Tr}(.)$ represents the trace operator. Hence, the first term of the optimization in (8) can be written as

$$\|Q - L^T X\|_F^2 = \text{Tr}((Q - L^T X)^T(Q - L^T X)). \quad (9)$$

Plugging $\mathbf{C}_\omega(X, \tau)$ and $\|Q - L^T X\|_F^2$ from Equations (6) and (9) into the optimization in (8) yields

$$X = \arg\min_X \text{Tr}\left((Q - L^T X)^T(Q - L^T X)\right)$$
$$+ \lambda \cdot \text{Tr}(A X^T X)$$
$$= \arg\min_X \text{Tr}\left(Q^T Q - 2Q^T L^T X + X^T L L^T X\right)$$
$$+ \lambda \cdot \text{Tr}(A X^T X)$$
$$\overset{(a)}{=} \arg\min_X \text{Tr}\left(-2Q^T L^T X + X^T L L^T X\right)$$
$$+ \lambda \cdot \text{Tr}(A X^T X)$$
$$= \arg\min_X \text{Tr}\left(C^T X\right) + \text{Tr}\left(D X X^T\right) + \lambda \cdot \text{Tr}\left(A X^T X\right), \quad (10)$$

**Algorithm 1** Manifold Extension for Time Series (METS)

**Inputs:** Embedding $\mathcal{L}$ for training points $\mathcal{U}$, parameters $K$ and $\alpha$, regularization parameter $\lambda$, temporal neighborhood graph $\mathcal{T}_K$ for extension set, squared distances $\Delta_X$ between training points $\mathcal{U}$ and extension points $\mathcal{Y}$, squared distances $\Delta_n$ between training points $\mathcal{U}$
**Outputs:** Embedding $\mathcal{X}$ for $\mathcal{Y}$
**Initialize:** $A \leftarrow \mathbf{0}_{N \times N}$
**for** $i = 1 \rightarrow N$ **do**
    **for** $j \in \mathcal{T}_K(y_i)$ **do**
        $B_{ij} \leftarrow \mathbf{0}_{N \times N}$
        $B_{ij}[i, i] \leftarrow 1, \ B_{ij}[j, j] \leftarrow 1$
        $B_{ij}[i, j] \leftarrow -1, \ B_{ij}[j, i] \leftarrow -1$
        $\omega_{ij} \leftarrow \exp(-\alpha \cdot \tau_{ij})$
        $A \leftarrow A + \omega_{ij} B_{ij}$ {Constructing weighted Laplacian matrix $A$ for temporal neighborhood graph $\mathcal{T}_K$}
    **end for**
**end for**
$D \leftarrow L L^T$
$Q \leftarrow \frac{1}{2}(\bar{\Delta}_n \mathbf{1}_N^T - \Delta_X)$
$C \leftarrow -2LQ$
$\text{vect}(X) \leftarrow -\frac{1}{2}\left((I \otimes D) + \lambda \cdot (A^T \otimes I)\right)^{-1} \cdot \text{vect}(C)$
$\mathcal{X} = \{x_1, x_2, \ldots, x_N\}$     {$x_i$ is the $i$-th column of X}
**return** $\mathcal{X}$

where in $(a)$ the constant matrix $Q^T Q$ has been dropped from the objective function. Note that we denote $C = -2LQ$ and $D = LL^T$. Additionally, we note that $\text{Tr}\left(X^T D X\right) = \text{Tr}\left(D X X^T\right)$ due to invariance of trace under cyclic permutations. Next, we denote the objective function in (10) by $J$, and take the derivative of $J$ with respect to the embedding matrix $X$ as follows:

$$\frac{\partial J}{\partial X} = C + (D + D^T)X + \lambda \cdot X(A + A^T)$$
$$\overset{(b)}{=} C + 2DX + 2\lambda X A, \quad (11)$$

where $(b)$ is due to the matrices $D$ and $A$ being symmetric. Solving $\frac{\partial J}{\partial X} = \mathbf{0}$ for $X$ gives us the solution to the optimization in (10), where $\mathbf{0}$ is an all-zeros matrix of size $N \times m$. In order to solve the matrix equation $\frac{\partial J}{\partial X} = \mathbf{0}$, we use the Kronecker product and the vectorized format of each term:

$$\text{vect}\left(\frac{\partial J}{\partial X}\right) = \text{vect}(C + 2DX + 2\lambda X A)$$
$$= \text{vect}(C) + 2 \ (I \otimes D) \ \text{vect}(X)$$
$$+ 2\lambda \ (A^T \otimes I) \ \text{vect}(X), \quad (12)$$

where $\otimes$ designates the Kronecker product operator and $I$ is the identity matrix. Setting (12) equal to zero and solving for $\text{vect}(X)$ yields:

$$\text{vect}(X) = -\frac{1}{2}\left((I \otimes D) + \lambda \cdot (A^T \otimes I)\right)^{-1} \cdot \text{vect}(C), \quad (13)$$

which provides the embedding obtained from the solution to (5). For brevity, we refer to the proposed algorithm as *manifold extension for time series* (METS), a summary of which is presented in Algorithm 1.

Note that Equation (13) involves a matrix inversion. Considering that $\text{rank}(A_1 \otimes A_2) = \text{rank}(A_1) \times \text{rank}(A_2)$, analysis of the rank of the first and second terms in the matrix under inversion boils down to that of $D$ and $A$, respectively. The matrix $D$ is invertible since the rows of $L$ are orthonormal. Since $A$ is a weighted Laplacian matrix, it is of rank $N-1$ with the all-ones vector being the eigenvector corresponding to the zero eigenvalue. However, the sum of $I \otimes D$ and $A^T \otimes I$ is full rank unless we have a pathological data structure.

Note also that the complexity of (13) is $\mathcal{O}(\max\{nmN, (mN)^\epsilon\})$, where $\mathcal{O}((mN)^\epsilon)$ is the complexity of matrix inversion for an $mN \times mN$ matrix. In comparison, the complexity of Isomap OoSE from Equation (2) is $\mathcal{O}(nmN)$. Hence, as long as the number of training data points $n$ is sufficiently large, compared to $m$ and $N$, the complexity of METS matches that of Isomap OoSE. Note that the above-mentioned computational complexities do not include the computation of the geodesic distances.

## IV. NUMERICAL EXPERIMENTS:
## OUT-OF-SAMPLE EXTENSION OF TIME SERIES

In this section we investigate the robustness of the proposed algorithm to different types of artifacts. In all the experiments, we consider image manifold datasets, where each data point corresponds with an image. We start from a dataset of original images that we treat as "clean" images, and we synthesize noisy images by adding several types of noise and artifacts to them. We consider three types of noise and artifacts: salt-and-pepper noise, Gaussian noise imposed on the pixel intensities, and motion blurring as caused by camera movements. Algorithms that are resilient to different noise/artifact types are highly desirable in practice as inferring noise types is often challenging. For our experiments, we consider two datasets: the eyeglasses dataset [14] and the Statue dataset [23].

The eyeglasses dataset is a custom eye-tracking dataset of $111 \times 112$-pixel captures from a computational eyeglasses prototype. The prototype features low-power cameras mounted on a pair of eyeglasses that are trained on the user's eyes, with the goal of tracking the gaze location of the user over time. The dataset contains $n + N = 900 + 100$ images and the dimensionality of the learned manifold is $m = 2$. A uniform sequential ordering exists among the $N$ out-of-sample points. A subset of consecutive out-of-sample points is shown in Figure 1a. Example noisy images with the three types of artifacts under study are also depicted in Figure 2.

The Statue dataset consists of 944 images of a rigid object on a turntable platform captured from a camera on an elevating arm. The images are captured every 6 degrees of rotation from 6 to 354 degrees and every 6 degrees of elevation from 0 to 90 degrees. Each image is cropped and resized to $128 \times 128$. In our experiments, we use half of the images at one of the elevation levels as out-of-sample points (i.e., $N = 29$ successive rotations in that elevation level form the out-of-sample points) and use the remaining points ($n = 915$) as training data. Note that this situation represents a practical scenario since we are testing over a missing slice of the manifold corresponding to the rotations in an elevation level of the camera. Sample images from the Statue dataset are shown in Figure 1b.

The experiments in this section pursue the following general framework. First, we obtain the Isomap embedding for $n$ training points. Next, we obtain the OoSE of both clean and noisy $N$ out-of-sample points. The OoSE of clean points is considered as a reference for performance measure, and we would like the embedding of the noisy version of the out-of-sample points to be as close to that of the clean data as possible. We set $k = 20$ in all the experiments for both the eyeglasses and Statue datasets. For each value of the noise parameter, we generate 6 instances of noise. We use the first instance of noise to find the best-performing value of the parameter $\lambda$ by a grid search. The selected value of $\lambda$ (which is different in general for different noise parameters) is used to measure the performance of the proposed algorithm on the remaining instances of noise. Finally, we average the error over the latter instances.

Note that for Isomap we cannot directly compare the two sets of out-of-sample points [24], as the embeddings learned from different samplings of the manifold are often subject to translation, rotation, and scaling. These variations must be addressed via manifold alignment before the embedded points are compared. We find the optimal alignment of the clean and noisy embeddings via *Procrustes analysis* [25], [26] and apply the resulting translational, rotational, and scaling components on the OoSE points. Finally, we measure the OoSE error $\mathcal{E}$ as the average $\ell_2$ distance between matching points in the two embeddings, i.e., out-of-sample extension of aligned clean and denoised testing data:

$$\mathcal{E} = \frac{1}{N} \sum_{i=1}^{N} \|z_i - w_i\|_2,$$

where $\mathcal{Z} = \{z_1, z_2, \ldots, z_N\}$ is the out-of-sample extension of clean data via Isomap OoSE, and $\mathcal{W} = \{w_1, w_2, \ldots, w_N\}$ is the OoSE from noisy data for the algorithm under test after alignment with $\mathcal{Z}$.

We compare the performance of the proposed algorithm against an adaptation of Isomap out-of-sample extension to ST-Isomap. We note that we compared the results of the proposed algorithm with those of ST-Isomap since it can be extended to out-of-samples in a similar fashion to Isomap OoSE. To obtain the out-of-sample extension of ST-Isomap, we find the distances of the out-of-sample points with training points via the neighborhood graph of the ST-Isomap algorithm, and then use (2) to find the embedding for the out-of-sample points. Note that since Isomap and ST-Isomap use different neighborhood graphs regardless of the parameter values of the ST-Isomap, the embedding of clean data via ST-Isomap differs from that of Isomap. Hence, we cannot directly compare the two embedding spaces. Thus we first align the two OoSE embedding spaces, and then evaluate the error between the aligned embedding obtained via ST-Isomap on clean and noisy data. We set the value of the parameter $c_{ATN} = 1$ as experimentally this selection produces the minimum error. We then perform a grid search over the window size $\epsilon$ and parameter $c_{CTN}$, in a similar manner to the evaluation of the parameter of the proposed algorithm, i.e., using 6 instances of noise. Note that we are giving an advantage to ST-Isomap out-of-sample extension over our proposed method by supplying temporal information about both training and out-of-sample points to the algorithm.

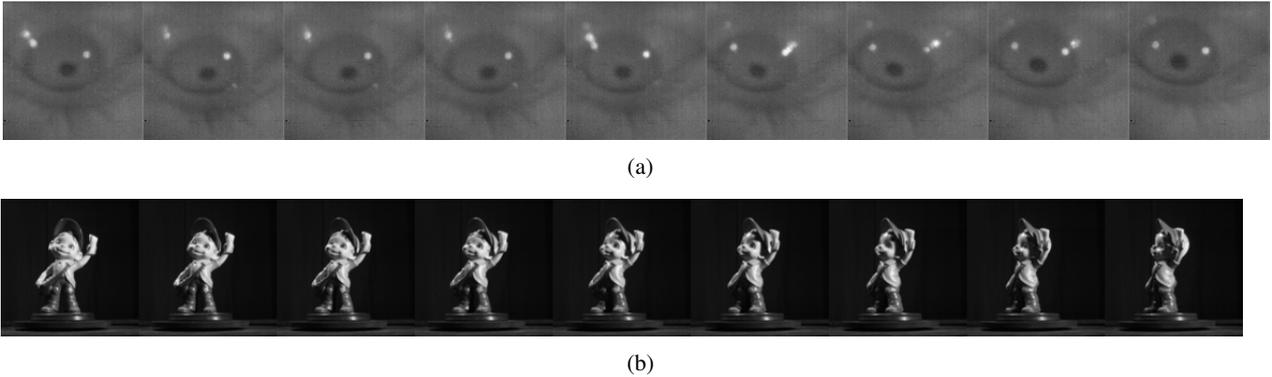

(a)

(b)

Fig. 1: Sample images used for out-of-sample extension. (*a*) Eyeglasses dataset; every second image in a sequence of 18 successive data points, (*b*) Statue dataset; 9 successive data points corresponding to successive rotations in an elevation level of the camera.

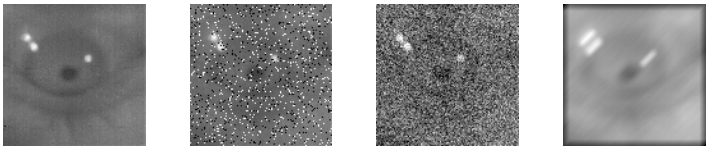

Fig. 2: Sample noisy images from the eyeglasses dataset. From left to right: noise-free, salt-and-pepper noise with $p = 0.2$, Gaussian noise with $\sigma = 0.1$, and motion blur with length of 20 pixels and angle of 45 degrees.

As mentioned in Section II, the algorithms on manifold denoising fail to address the out-of-sample extension problem. In addition, none of the mentioned algorithms take advantage of temporal information among the images. In order to make a fair comparison with manifold denoising algorithms (as baselines), we apply manifold denoising algorithms on the collection of both training and out-of-sample points, and compare the embedding for the denoised version of out-of-sample points with that of the clean data. Since the manifold denoising algorithms take advantage of the neighboring points to obtain the denoised version of each point, the inclusion of clean points during denoising improves the quality of denoising for the (noisy) out-of-sample points.

We consider MBMS, which is one of the most recent algorithms for denoising of manifolds. We also considered LLD, an alternative manifold denoising algorithm, but in our experiments it turned out to be less resilient to noise. Specifically, for the range of noise levels that we consider in our experiments, LLD failed to perform well and the embedding of the resulting denoised points was comparable to that of the noisy points. This can be associated with the neighborhood selection mechanism employed in LLD where, for each point, the set of points that include that point in their neighborhood is considered during denoising. Since at higher noise levels the majority of neighborhoods for clean points is composed of other clean points, the noisy points mostly belong to the neighborhoods of noisy points. Hence, the denoised version of the noisy point is not an improvement over the noisy point. In contrast, the MBMS algorithm denoises each point by considering its neighboring points, i.e., MBMS considers neighbors in the opposite direction to that of LLD. It turns out that the neighborhood of a noisy point includes a majority of clean points and a minority of noisy points. Hence, the denoised version of that point is a major improvement over the counterpart noisy version. As such, we do not consider LLD and only include MBMS in our experimental results.

For MBMS, we set the value of the parameters $L = 1$ or $L = 2$ (whichever provides the best performance) and the number of iterations to 3, as suggested by the authors; our experiments also confirms that this setting produces the best result. In order to choose the value of the parameter $\sigma$, we perform a grid search over the interval $[1, 20]$ on the first instance of noise, evaluate the algorithm with the chosen value of $\sigma$ on the remaining instances, and report the average error.

### A. Salt-and-Pepper Noise

We consider salt-and-pepper noise where the intensity of each pixel is randomly flipped with a probability of $p$

to either zero or one, with both having equal probability $\frac{p}{2}$. We vary $p$ from 0.2 to 0.6 with a step size of 0.1. For the eyeglasses dataset, the selected values for the parameter $\lambda$ are $[0.02, 0.14, 0.2, 0.53, 9.0] \times 10^5$, respectively. Figure 4a compares the performance of the proposed algorithm at different noise levels against that of Isomap OoSE, ST-Isomap OoSE, and MBMS, for the eyeglasses dataset. As can be observed from this figure, the average $\ell_2$-norm error of the proposed algorithm is lower than those of Isomap OoSE and ST-Isomap OoSE. The error bars indicate the variability over the five noisy instances of the out-of-sample points via the standard error of the mean (SEM). Note that SEM is computed as $\frac{s}{\sqrt{5}}$, where $s$ is the sample standard deviation. We can observe from the plots that the variability of all the algorithms, specifically METS, is quite small for both salt-and-pepper and Gaussian noises. We see a higher variability for the motion blur artifact experiments, which is to be expected since the latter artifact model is not zero-mean and has a higher variation over the pixels (due to rotations).

Recall that METS incorporates two parameters: neighborhood size $K$ of the temporal graph and decay parameter $\alpha$ of the time weighting function. Figure 3 demonstrates the performance of METS at different values of the parameters $K$ and $\alpha$ for a time series contaminated by an instance of salt and pepper noise with $p = 0.4$. As we can see from this figure, METS achieves its best performance over a large range of the parameters $K$ and $\alpha$. Intuitively, either $K$ or $\alpha$ can be leveraged to set the level of dependence among the temporal neighbors in the embedding space. Experimentally, the observed pattern turned out to be similar at different noise levels; hence the values of the optimal parameters $K$ and $\alpha$ are fairly robust to noise level and are merely data dependent. Finally, we note that other choices for the weighting function are possible, and the performance of METS turned out not to be sensitive to the choice of the weighting function. As an example, we also tried a Gaussian kernel for the weighting function, and obtained virtually identical results to those for the exponential decay kernel. Intuitively, the Gaussian kernel behaves very similar to the exponential decay function in that it models the decrease in likelihood for the similarity of the embedding of each point with its subsequent temporal neighbors; the only difference is that the Gaussian functions decay with respect to time is faster than that of the exponential function.

As a rule of thumb, we may follow one of the following procedures in selection of the parameters $K$ and $\alpha$: ($i$) pick a large value for the parameter $K$ (it can be set to $N$, in which case we essentially drop the parameter $K$ from the algorithm) and tune the value of the parameter $\alpha$, or ($ii$) set $\alpha$ to a small value (it can be set to zero, in which case we essentially drop the parameter $\alpha$ from the algorithm) and tune the value of the parameter $K$. The equivalence of the two procedures can be explained as follows. Assume that $K_1$ is the optimal neighborhood size in the procedure ($i$), which indicates the optimal number of temporal neighbors to be incorporated in the embedding. Intuitively, the excess of the temporal neighbors $K_2 > K_1$ in procedure ($ii$) can be compensated by increasing the decay parameter $\alpha$ so that the impact of the extra neighbors is mostly ignored in the embedding. Throughout the experiments we follow the second procedure. For the eyeglasses dataset, we set $K = 10$ and $\alpha = 0$ (alternatively, we can set $K = 20$ and $\alpha = 0.3$ and virtually achieve the same result). In the experiments for the Statue dataset, we set $K = 2$ and $\alpha = 0$.

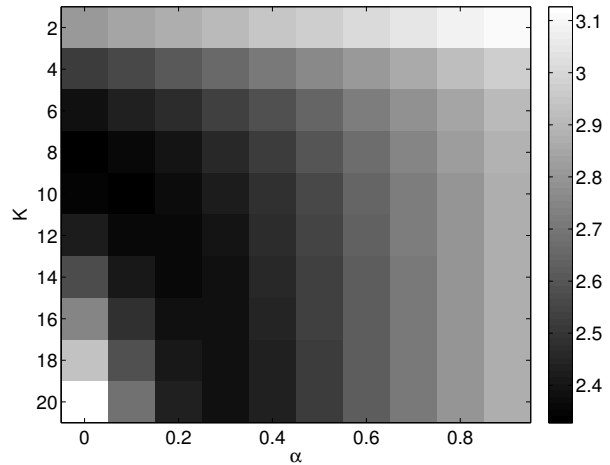

Fig. 3: Evaluation of METS at different values for the parameters $K$ and $\alpha$.

For the Statue dataset, the selected values of the parameter $\lambda$ for METS are $[0.5, 1.5, 3, 9.5, 12] \times 10^6$ at different noise levels. Figure 4d compares the performance of METS at different noise levels against that of Isomap OoSE, ST-Isomap OoSE, and MBMS, for the Statue dataset. As we can see from this figure, MBMS only improves the performance marginally over the plain Isomap OoSE. Moreover, ST-Isomap genrally does not perform well for the Statue dataset.

### B. Gaussian Noise

We consider Gaussian noise with zero mean and variance of $\sigma$, where we vary $\sigma$ from 0.1 to 0.5 with the step size of 0.1. When using the eyeglasses dataset, the selected values for the parameter $\lambda$ are $[0.25, 1.25, 1.75, 2.0, 4.5] \times 10^4$, respectively. Figure 4b compares the performance of the proposed algorithm at different Gaussian noise variance levels against that of Isomap OoSE, ST-Isomap OoSE, and MBMS.

Figure 5a depicts the Isomap trajectory for the clean out-of-sample points. Its noisy counterpart with the addition of an instance of Gaussian noise with $p = 0.2$ is shown in Figure 5b. In each plot, sequentially adjacent points are connected by a blue line and temporal order is color coded from blue to red. We obtained the out-of-sample extension of the noisy data by the proposed algorithm with three settings of the parameter $\lambda$; $\lambda = 0.5 \times 10^4$, $1.25 \times 10^4$, and $10 \times 10^4$ as depicted in Figures 5d, 5e, and 5f, respectively. Among the mentioned values of the parameter $\lambda$, $\lambda = 1.25 \times 10^4$ produces the minimum $\ell_2$-norm error in the recovered trajectory when compared to Isomap trajectory for the clean out-of-sample points. This can be verified by the similarity of the trajectory shown in Figure 5e to that of Figure 5a. As can be observed from the figure, a smaller $\lambda$ overfits the noisy trajectory and retains the noise in the embedding, whereas a larger one oversmooths the trajectory. Increasing the value of the parameter $\lambda$ eventually converts the embedding's trajectory to an almost-straight curve. This is to be expected since eventually the spatial information will be neglected and the embedding will retain only the temporal information of the time series. Note that setting $\lambda = 0$ in the proposed algorithm, converts it into the plain Isomap OoSE and returns the noisy trajectory, i.e., Figure 5b.

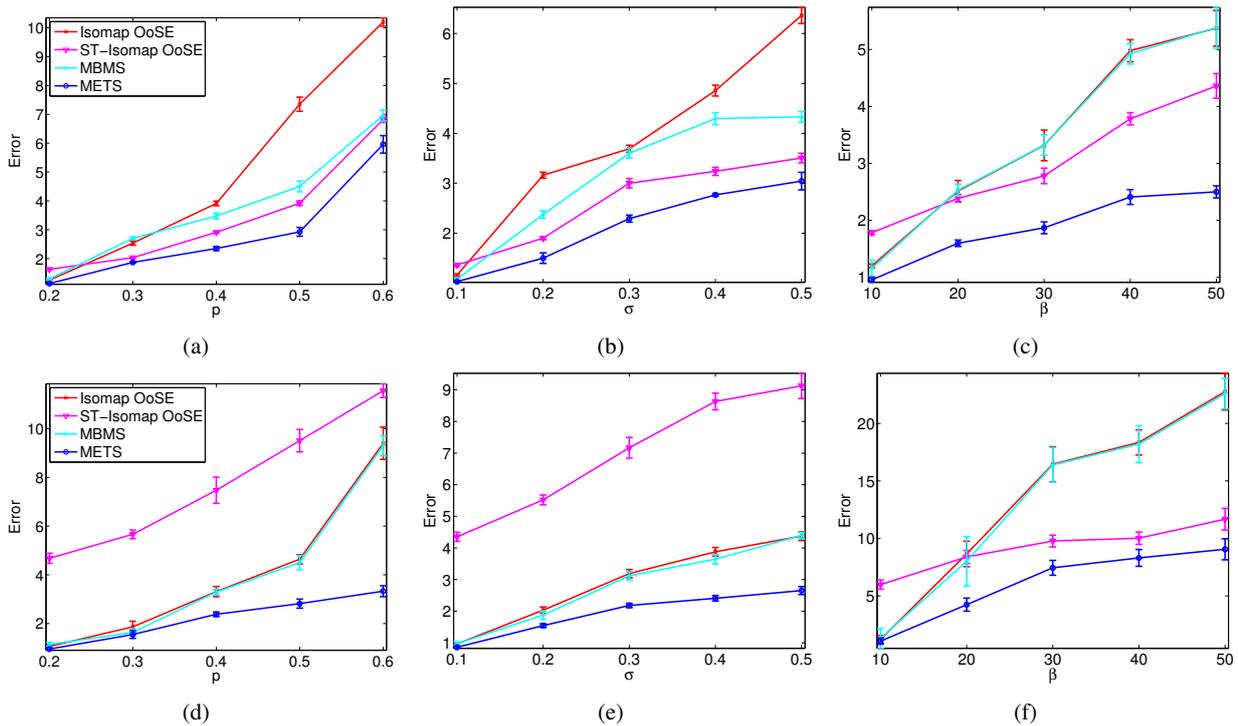

Fig. 4: Performance evaluation for out-of-sample extension of the data points with salt and pepper noise (first column), Gaussian noise (second column), and motion blur artifact (third column) for eyeglasses dataset (first row), and Statue dataset (second row).

Figure 5c shows the denoised trajectory obtained via MBMS. Figure 6 demonstrates the aligned ST-Isomap OoSE trajectory on clean and noisy data. We note that for the other types of noise considered in the sequel, the embeddings obtained from the different algorithms we test are reminiscent of those shown in Figure 5 and 6, and thus we do not include them in this paper.

For the Statue dataset, the selected values of the parameter $\lambda$ are $[0.1, 1.3, 2.3, 4.4, 7] \times 10^6$ at different noise levels. Figure 4e demonstrates the performance of different algorithms at different noise levels, which is similar to those for the salt and pepper noise for the same dataset.

### C. Motion Blur

We use a linear motion model to simulate the motion blur artifact. The artifact is applied via convolution by a filter that approximates the linear motion of a camera by $\eta$ pixels, with an angle of $\theta$ degrees in a counterclockwise direction. We consider a uniform distribution for $\theta$ over the interval $[0°, 360°]$ and a Gamma distribution for $\eta$ (with PDF of $p(x) = \frac{1}{\Gamma(\alpha)\beta^\alpha} x^{\alpha-1} \exp(\frac{-x}{\beta})$, where $\Gamma(.)$ denotes the Gamma function). We set the shape parameter of the Gamma PDF $\alpha = 1$ and vary the scale parameter $\beta$ from 10 to 50 by step size of 10 in order to produce different strengths of motion blur. The selected values for the parameter $\lambda$ are $[0.2, 1.0, 1.3, 1.6, 2.4] \times 10^4$, respectively. Figure 4c compares the performance of the proposed algorithm at different values of the scale parameter $\beta$ against that of Isomap OoSE, ST-Isomap OoSE, and MBMS.

For the Statue dataset, the following values of the parameter $\lambda$ are selected when we vary $\beta$ from 10 to 50, respectively: $[0.3, 9, 11, 15, 100] \times 10^6$. Figure 4f compares the performance of different algorithms on the Statue dataset contaminated by the motion blur artifact. From the figure, ST-Isomap outperforms Isomap OoSE and MBMS at noise levels above $\beta = 20$. As for other artifacts, METS outperforms Isomap OoSE, MBMS, and ST-Isomap.

## V. NUMERICAL EXPERIMENTS: APPLICATION IN EYE GAZE ESTIMATION

Finally, we consider an application of manifold models in our motivating computational eyeglasses platform. More precisely, we focus on the Eyeglasses dataset, illustrated in Figure 1a, which is collected for the purpose of training an estimation algorithm for eye gaze position in a 2-D image plane. The dataset corresponds to a collection of image captures of an eye from a camera mounted on an eyeglass frame as the subject focuses their gaze into a dense grid of known positions (size $111 \times 112$, covering a 36 degrees of arc) that is used as ground truth.

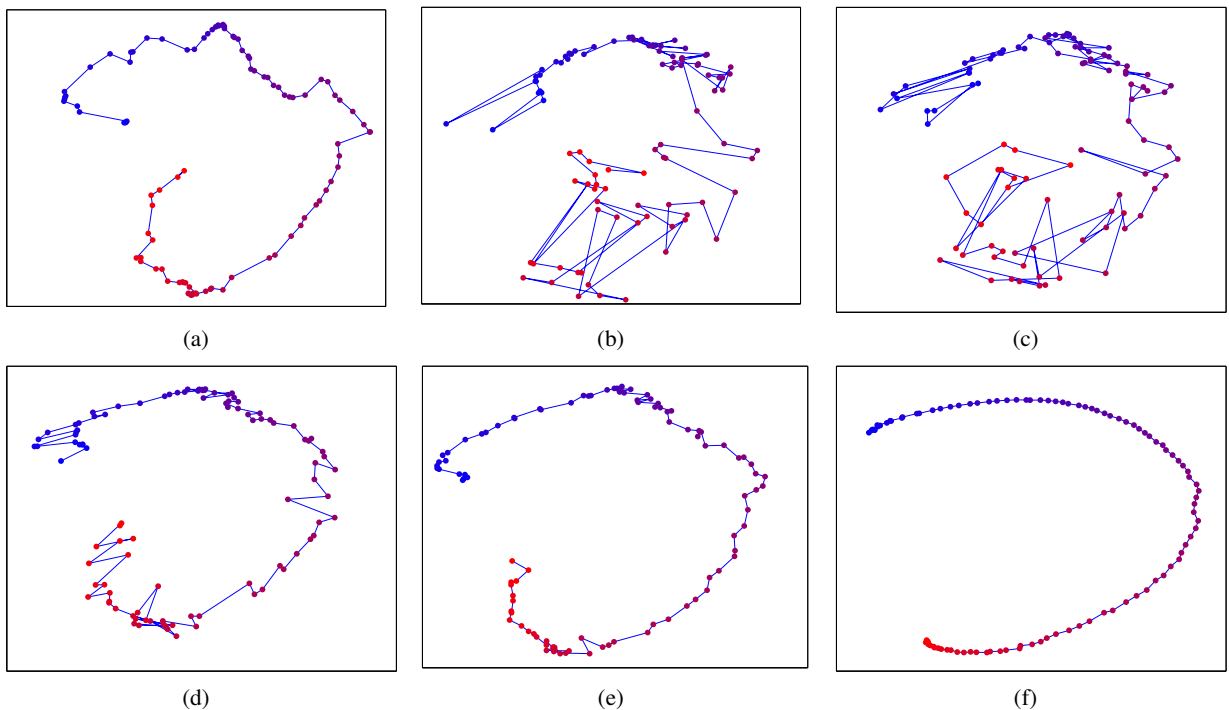

(a)       (b)       (c)

(d)       (e)       (f)

Fig. 5: Effect of the value of the parameter $\lambda$ on the embedding of out-of-sample points for the eyeglasses dataset. ($a$) Isomap OoSE trajectory for clean data, ($b$) Isomap OoSE trajectory for noisy data with Gaussian noise with $\sigma = 0.2$, ($c$) OoSE trajectory of data denoised via MBMS, OoSE of noisy data via METS for ($d$) $\lambda = 0.5 \times 10^5$; ($e$) $\lambda = 1.25 \times 10^5$, and ($f$) $\lambda = 10 \times 10^5$, respectively.

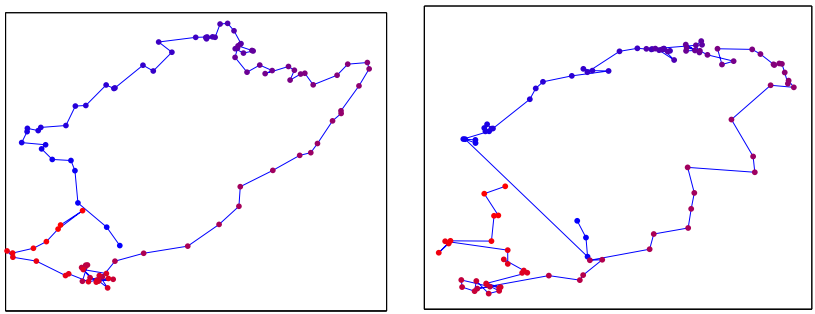

Fig. 6: ST-Isomap OoSE trajectory of the eyeglasses dataset for ($a$) clean and ($b$) noisy data with Gaussian noise with $\sigma = 0.2$.

Two primary approaches has been used in the literature for eye gaze estimation. In *feature-based* approaches, geometric features such as contours and corners of the pupil, limbus and iris, are used to extract features of the eye [27], [28]. The drawback of this approach is that it works best with Near-Infrared (NIR) illumination sources, which, through their positioning relative to the camera, can make the pupil appear brighter or darker, thereby making it easier to detect the boundary [14]. When using visible light, feature-based techniques are harder to use since the boundary of the pupil is harder to detect.

Alternatively, *appearance-based* tracking algorithms attempt to predict the gaze location directly from the pixels of the eye image without an intermediate geometric representation of the pupil. Two prominent appearance-based methods used in the gaze tracking literature are multilayer Perceptron (MLP) [14], [29] and manifold embeddings [30].

The idea behind the manifold-based method is to find a nonlinear manifold embedding of the original dataset $\mathcal{X}$ and extend it to the 2-D parameter space samples given by the eye gaze ground truth. The proposed method employs the weights obtained by LLE, when applied to the training image dataset together with a testing image $\mathcal{X} \cup \{x_t\}$, and applies these weights in the parameter space to estimate the parameters of the test point.

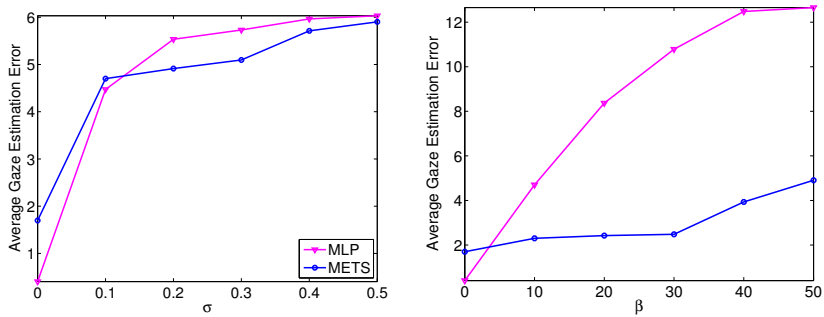

Fig. 7: Eye gaze estimation result for noisy time series contaminated with (left) Gaussian noise and (right) motion blur artifact.

We evaluate the performance of METS on eye gaze estimation for a time series contaminated by Gaussian and motion blur artifacts. We compare the performance of METS against that of MLP. Note that for each point in the time series the LLE weights used in the manifold-based eye gaze estimation are set in terms of the training data points.

In a similar fashion to the previous section, we generate 6 instances of the noisy time series. The first instance is used to determine the values of the parameter $\lambda$ via a grid search; we then apply the selected values of the parameter lambda for the remaining instances of the noisy time series and average the performance over those instances. The same procedure is used to set the number of hidden units of the MLP. Note that for the MLP model we follow the procedure used in [14], though we remove the penalty factor that enforces image masking (i.e., selecting a subset of the pixels from the eye images) by setting the value of the regularization parameter to zero. Experimentally, we observed that removing this restriction improves the performance of the MLP model in eye gaze estimation.

Figure 7 shows the average eye gaze estimation error $e$ (in terms of degrees of arc) as a function of the noise parameter for time series contaminated with Gaussian noise and motion blur artifact. As can be observed from the plots, the MLP model outperforms the manifold-based approach in the noise-free case. More specifically, the MLP model estimates the eye gaze estimation within $0.46$ degrees of the true values on average, whereas the manifold-based approach incurs an error of $1.69$ degrees. In the case of the Gaussian noise, the two methods perform comparable to each other, with METS outperforming the MLP model as we increase the noise level. However, in the case of the motion blur artifact, METS quickly outperforms the MLP model by a large margin as we add noise to the time series.

The poor performance of the MLP model for the motion blur artifact can be associated with its systematic impact that does not average out in hidden units. On the contrary, the time series contaminated by the Gaussian noise is zero-mean and locally averages out in hidden units. In other words, each hidden unit is activated by a weighted combination of all the pixels of an input image. The value of this combination for the random perturbation incurred by the Gaussian noise cancels out in expectation. While the MLP model performs poorly for some artifact models, our proposed algorithm shows more resilience to a wider class of noise models and outperforms the MLP model by a large margin for systematic artifact models such as the motion blur artifact.

## VI. Discussion

In this paper we devised an extension of Isomap for sequentially ordered out-of-sample data points. Numerical experiments indicate the robustness of the proposed algorithm against different noise and artifact models. The smoothness/behavior of the embedding is determined by the regularization parameter $\lambda$, the optimal value of which depends on the level of noise/artifacts. The more noise is present in the data, the larger the value of the parameter $\lambda$ needs to be in order to recover the embedding of the clean out-of-sample data. Future work can address automatic selection of the regularization parameter as a function of the noise parameters.

In the future, we will evaluate the application of the proposed algorithm on object tracking problems. In addition, it would be interesting to explore the possibility of employing our proposed spatio-temporal compactness to other manifold learning algorithms such as LLE. We also expect our formulation of spatio-temporal compactness to leverage other types of correlation present among the data points. As an example, manifold-based hyperspectral unmixing fits a manifold model to nonlinear mixtures of endmember spectra. It is expected that mixtures corresponding to neighboring pixels in a hyperspectral image will feature highly correlated mixing abundances, leading to a desired correlation of distances between pixels and geodesic distances between the corresponding spectra as points in the manifold.

## VII. Acknowledgements

We thank Addison Mayberry for providing us with the eyeglasses dataset and the codes for the implementation of the MLP eye gaze estimation algorithm.

## References

[1] H. Dadkhahi, M. F. Duarte, and B. Marlin, "Isomap out-of-sample extension for noisy time series data," in *IEEE Int. Workshop on Machine Learning for Signal Processing (MLSP)*, Boston, MA, Sept. 2015, pp. 1–6.

[2] C. M. Bishop, *Pattern Recognition and Machine Learning (Information Science and Statistics)*. Secaucus, NJ: Springer-Verlag, 2006.

[3] J. B. Tenenbaum, V. d. Silva, and J. C. Langford, "A global geometric framework for nonlinear dimensionality reduction," *Science*, vol. 290, no. 5500, pp. 2319–2323, 2000.

[4] S. T. Roweis and L. K. Saul, "Nonlinear dimensionality reduction by locally linear embedding," *Science*, vol. 290, no. 5500, pp. 2323–2326, 2000.

[5] M. Belkin and P. Niyogi, "Laplacian eigenmaps for dimensionality reduction and data representation," *Neural Computation*, vol. 15, no. 6, pp. 1373–1396, Mar. 2003.

[6] D. L. Donoho and C. Grimes, "Hessian eigenmaps: Locally linear embedding techniques for high-dimensional data," *Proc. Nat. Acad. Sciences (PNAS)*, vol. 100, no. 10, pp. 5591–5596, May 2003.

[7] O. C. Jenkins and M. J. Matarić, "A spatio-temporal extension to Isomap nonlinear dimension reduction," in *Int. Conf. on Machine Learning (ICML)*, 2004, pp. 441–448.

[8] M. Lewandowski, J. Martinez-del Rincon, D. Makris, and J. C. Nebel, "Temporal extension of Laplacian eigenmaps for unsupervised dimensionality reduction of time series," in *Int. Conf. on Pattern Recognition (ICPR)*, Washington, DC, USA, 2010, pp. 161–164.

[9] R. S. Lin, C. B. Liu, M. H. Yang, N. Ahuja, and S. E. Levinson, "Learning nonlinear manifolds from time series," in *European Conf. on Computer Vision (ECCV)*, Graz, Austria, 2006, pp. 245–256.

[10] S. Roweis, L. K. Saul, and G. E. Hinton, "Global coordination of local linear models," in *Neural Info. Proc. Systems (NIPS)*, Vancouver, BC, 2002, pp. 889–896.

[11] A. Rahimi, B. Recht, and T. Darrell, "Learning appearance manifolds from video," in *Computer Vision and Pattern Recognition (CVPR)*, San Diego, CA, USA, 2005, pp. 868–875.

[12] Z. Zhao and A. Elgammal, "Human activity recognition from frame's spatiotemporal representation," in *Int. Conf. on Pattern Recognition (ICPR)*, Tampa, Florida, 2008, pp. 1–4.

[13] M. Balasubramanian and E. L. Schwartz, "The Isomap algorithm and topological stability," *Science*, vol. 295, no. 5552, p. 7, 2002.

[14] A. Mayberry, P. Hu, B. Marlin, C. Salthouse, and D. Ganesan, "iShadow: Design of a wearable, real-time mobile gaze tracker," in *Int. Conf. Mobile Systems, Applications and Services (MobiSys)*, Bretton Woods, NH, June 2014, pp. 82–94.

[15] T. Cox and M. Cox, *Multidimensional scaling*. Boca Raton : Chapman & Hall/CRC, 2001.

[16] V. d. Silva and J. B. Tenenbaum, "Global versus local methods in nonlinear dimensionality reduction," in *Neural Info. Proc. Systems (NIPS)*, Vancouver, BC, Dec. 2003, pp. 705–712.

[17] Y. Bengio, J.-F. Paiement, and P. Vincent, "Out-of-sample extensions for LLE, Isomap, MDS, Eigenmaps, and spectral clustering," in *Neural Info. Proc. Systems (NIPS)*, Vancouver, Canada, 2003, pp. 177–184.

[18] H. Chen, G. Jiang, and K. Yoshihira, "Robust nonlinear dimensionality reduction for manifold learning," in *Int. Conf. on Pattern Recognition (ICPR)*, vol. 2, Hong Kong, 2006, pp. 447–450.

[19] M. Hein and M. Maier, "Manifold denoising," in *Neural Info. Proc. Systems (NIPS)*, Vancouver, Canada, 2006, pp. 561–568.

[20] Y. Gao, K. L. Chan, and W. y. Yau, "Manifold denoising with Gaussian process latent variable models," in *Int. Conf. on Pattern Recognition (ICPR)*, Tampa, FL, Dec. 2008.

[21] D. Gong, F. Sha, and G. Medioni, "Locally linear denoising on image manifolds," in *Artificial Intelligence and Statistics (AISTATS)*, Sardinia, Italy, 2010, pp. 265–272.

[22] W. Wang and M. Á. Carreira-Perpin, "Manifold blurring mean shift algorithms for manifold denoising." in *Computer Vision and Pattern Recognition (CVPR)*, San Francisco, CA, 2010, pp. 1759–1766.

[23] R. Pless and I. Simon, "Using thousands of images of an object," in *Int. Conf. on Computer Vision, Pattern Recognition and Image Processing*, Research Triangle Park, NC, 2002, pp. 684–687.

[24] F. Dornaika and B. Raduncanu, "Out-of-sample embedding for manifold learning applied to face recognition," in *IEEE Conf. Computer Vision and Pattern Recognition Workshops (CVPRW)*, Portland, OR, June 2013, pp. 862–868.

[25] G. H. Golub and C. F. Van Loan, *Matrix Computations (3rd Ed.)*. Baltimore, MD: Johns Hopkins University Press, 1996.

[26] C. Wang and S. Mahadevan, "Manifold alignment using procrustes analysis," in *Int. Conf. Machine Learning (ICML)*, New York, NY, 2008, pp. 1120–1127.

[27] C. Morimoto and M. Mimica, "Eye gaze tracking techniques for interactive applications," *Computer Vision and Image Understanding*, vol. 98, no. 1, pp. 4–24, Apr. 2005.

[28] D. W. Hansen and Q. Ji, "In the eye of the beholder: A survey of models for eyes and gaze," *IEEE Trans. Pattern Analysis and Machine Intelligence*, vol. 32, no. 3, pp. 478–500, Mar. 2010.

[29] S. Baluja and D. Pomerleau, "Non-intrusive gaze tracking using artificial neural networks," in *Advances in Neural Information Processing Systems (NIPS)*, Pittsburgh, PA, 1994, pp. 753–760.

[30] K. Tan, D. Kriegman, and N. Ahuja, "Appearance-based eye gaze estimation," in *IEEE Workshop on the Application of Computer Vision (WACV)*, Orlando, FL, Dec. 2002, pp. 191–195.